# A Structured Reasoning Framework for Unbalanced Data Classification Using Probabilistic Models

Junliang Du
Shanghai Jiao Tong University
Shanghai, China

Shiyu Dou
Yale University
New Haven, USA

Bohuan Yang
University of California San Diego
San Diego, USA

Jiacheng Hu
Tulane University
New Orleans, USA

Tai An*
University of Rochester
Rochester, China

*Abstract*-This paper studies a Markov network model for unbalanced data, aiming to solve the problems of classification bias and insufficient minority class recognition ability of traditional machine learning models in environments with uneven class distribution. By constructing joint probability distribution and conditional dependency, the model can achieve global modeling and reasoning optimization of sample categories. The study introduced marginal probability estimation and weighted loss optimization strategies, combined with regularization constraints and structured reasoning methods, effectively improving the generalization ability and robustness of the model. In the experimental stage, a real credit card fraud detection dataset was selected and compared with models such as logistic regression, support vector machine, random forest and XGBoost. The experimental results show that the Markov network performs well in indicators such as weighted accuracy, F1 score, and AUC-ROC, significantly outperforming traditional classification models, demonstrating its strong decision-making ability and applicability in unbalanced data scenarios. Future research can focus on efficient model training, structural optimization, and deep learning integration in large-scale unbalanced data environments and promote its wide application in practical applications such as financial risk control, medical diagnosis, and intelligent monitoring.

I. INTRODUCTION

With the rapid development of data-driven technology, the existence of unbalanced data has become a common phenomenon in actual data mining and machine learning applications. Unbalanced data refers to the situation where there is a significant imbalance in the distribution of data of different categories, which is usually manifested as the amount of data in some categories is much larger than that in other categories. This asymmetry of data distribution is very common in many practical scenarios, such as rare disease detection in medical diagnosis [1], fraud detection in financial risk control [2-4], human-computer interaction gesture detection [5], etc. How to build accurate, efficient, and stable models in this environment has always been an important research topic in the field of data mining and machine learning.

As a powerful probabilistic graphical model, Markov network [6] can effectively represent the conditional dependency between random variables and is widely used in computer vision, natural language processing, and bioinformatics [7]. However, in applications facing unbalanced data, traditional Markov network methods often face multiple challenges. First, minority class samples in unbalanced data usually contain key decision information, but due to the small amount of data, the model tends to ignore this information during the learning process, resulting in classification biased towards the majority class. Second, the parameter estimation of Markov network depends on large-scale sample data. The imbalance of sample size will directly affect the accuracy of network structure and parameters, thereby reducing the generalization ability and prediction performance of the model [8].

Studies have shown that processing unbalanced data requires adjustments in multiple links such as model construction, feature selection, and learning algorithms. Specifically, the structural learning and parameter estimation of Markov networks should take into account the characteristics of unbalanced data, and adopt more effective marginal probability estimation and marginalization inference methods to alleviate the impact of uneven data distribution on model learning. In addition, data preprocessing methods based on resampling strategies (such as oversampling and undersampling), combined with weighted loss functions and cost-sensitive learning, can improve the model's recognition ability for minority class samples to a certain extent. By introducing latent variable modeling and structural optimization techniques, the adaptability and expressiveness of Markov networks in unbalanced data environments can be enhanced [9].

In terms of algorithm design, some enhanced Markov network algorithms have been proposed, such as the maximum entropy Markov model using adaptive weight learning and conditional random field variants for classification tasks. These improved methods improve the performance of the model in an unbalanced class environment by optimizing the objective function and introducing prior knowledge. In addition, the joint learning method of the generative model and the discriminative model based on the Markov network has gradually become a research hotspot, further promoting its application in unbalanced data analysis.

## II. RELATED WORK

Research on addressing unbalanced data in machine learning has seen significant advancements in recent years, particularly in the development of probabilistic graphical models, data balancing techniques, and deep learning frameworks. These works contribute directly to the challenges tackled in this paper, specifically the classification bias and insufficient minority class recognition associated with unbalanced data scenarios, and they provide insights into optimizing Markov networks for robust decision-making.

Jiang et al. [10] introduced generative adversarial networks (GANs) to balance class distributions by generating synthetic samples in financial market supervision tasks, showing improved performance in identifying minority classes. This concept of generating additional synthetic samples or weighting minority classes aligns with our use of weighted loss functions and probabilistic marginal estimation in Markov networks to better capture minority class behavior and enhance generalization. Similarly, Feng et al. [11] proposed a collaborative optimization strategy using deep learning models like ResNeXt to improve data mining tasks in financial datasets, emphasizing efficient feature extraction and model robustness. Their work highlights the benefit of combining multiple learning approaches, a principle reflected in our integration of structured probabilistic reasoning with optimization techniques.

Another key aspect of this paper is its focus on structural learning within Markov networks, which is closely related to research on graph-based models. Zhang et al. [12] explored robust graph neural networks (GNNs) for analyzing the stability of dynamic networks, demonstrating how structural adaptations improve model robustness in evolving data settings. This is conceptually related to our emphasis on structural dependency learning within Markov networks, which enhances the network's ability to reason globally over sparse and imbalanced data.

Several studies have also investigated adaptive parameter optimization for models trained on skewed datasets. Yan et al. [13] demonstrated how neural architecture search (NAS) can identify optimal configurations for deep models, highlighting the benefits of adaptive learning mechanisms in dynamic environments. In this paper, we similarly aim to extend Markov networks by exploring adaptive weight adjustments and structural optimization to balance decision-making between majority and minority classes. Furthermore, Huang et al. [14] applied reinforcement learning-based Q-learning for data mining in dynamic environments, showcasing adaptive optimization through reward-based strategies. This directly connects to our future work on integrating reinforcement learning with probabilistic reasoning to improve Markov networks' performance in large-scale and dynamic applications.

Deep learning has further facilitated advances in time-series analysis and high-dimensional data mining, which are relevant to the scalable design of Markov networks. Sun et al. [15] proposed Transformer models for time-series risk prediction, highlighting how sequential dependencies and temporal patterns can be effectively captured in prediction tasks. We draw inspiration from this work in our exploration of extending Markov networks to handle time-evolving data distributions, particularly when imbalanced classes vary over time.

Moreover, handling high-dimensional and sparse data efficiently is another important consideration for this study. Wang [16] proposed the use of sparse decomposition and adaptive weighting to effectively mine multimodal datasets, a technique that could complement our marginal probability estimation strategy by reducing noise and focusing on key features. Similarly, Li [17] discussed machine learning methods for recognizing patterns in high-dimensional data mining, which are applicable to our approach of capturing significant relationships among variables through Markov network structure learning.

The reviewed studies highlight key advancements in probabilistic modeling, adaptive optimization, and data augmentation techniques that have directly influenced the contributions of this paper. By integrating structured probabilistic reasoning, marginal probability estimation, and adaptive weight optimization, our proposed Markov network achieves robust classification in unbalanced data environments, demonstrating significant improvements over traditional models such as logistic regression, support vector machines, and random forests.

## III. METHOD

In the study of Markov networks for unbalanced data, the construction and reasoning of the model mainly revolve around the probabilistic dependency between nodes. A Markov network is an undirected graph model in which each node represents a random variable and the edges represent the mutual dependency between variables. The goal of the model is to model these random variables through joint probability distribution, reflecting the interaction and conditional independence between them. The model architecture is shown in Figure 1.

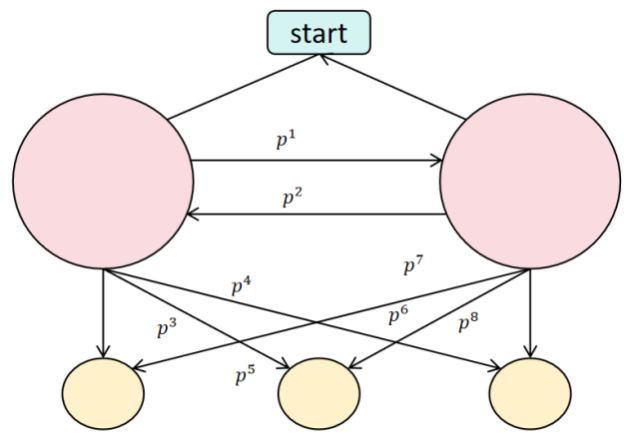

Figure 1 Overall model architecture

The definition of joint probability distribution is the core of Markov network. For an undirected graph $G = (V, E)$ containing a set of nodes V and a set of edges E, the joint probability distribution $P(X)$ can be decomposed into the product of potential functions through the maximum clique in the graph. The joint probability distribution form is expressed as:

$$P(X) = \frac{1}{Z} \prod_{C \in c} \phi_C(X_C)$$

When dealing with unbalanced data, it is necessary to enhance the learning ability of the model through marginal probability distribution estimation and conditional reasoning. The calculation of marginal probability distribution involves marginalizing the joint distribution, while the reasoning process requires estimating the conditional probabilities of certain nodes. In practical applications, Markov chain Monte Carlo (MCMC) sampling or variational reasoning is often used to approximate these probability calculations and reduce the computational complexity in high-dimensional space.

To address the class imbalance problem of unbalanced data, a weight adjustment mechanism can be introduced in the parameter learning stage of the model. By applying a cost-sensitive loss function to the objective function, the model's neglect of minority class samples can be effectively alleviated. The commonly used log-likelihood maximization objective function is as follows:

$$L(\theta) = \sum_{i=1}^{N} w_i \log(PX_i \mid \theta)$$

Among them, $w_i$ is the weight of sample $X_i$, reflecting its importance in the training process, and $\theta$ is the parameter set of the model. For minority class samples, higher weights can be given to improve the category recognition ability of the model and reduce the cost of misclassification.

In addition, in the structural learning stage, the edge connections of the Markov network can be optimized by introducing a regularization strategy of the graph structure to avoid the generalization ability of the model being reduced due to an overly complex graph structure. By maximizing the logarithmic posterior probability (MAP), the regularization term $R(\theta)$ is introduced in the structural learning.

$$J(\theta) = \sum_{i=1}^{N} w_i \log(P(X_i \mid \theta)) - \lambda R(\theta)$$

Among them, $\lambda$ is the regularization coefficient, which is used to balance the fit and complexity of the model. Through this formula, the model can achieve an effective trade-off between accuracy and complexity, and improve the robustness and generalization ability of the model in an unbalanced data environment.

In summary, the construction of Markov networks for unbalanced data requires a comprehensive design in terms of joint probability distribution modeling, parameter learning, and structural optimization. By introducing weight adjustment, cost-sensitive learning, and regularization strategies, the stability and performance of the model when processing unbalanced data can be enhanced, providing strong technical support for classification, prediction, and decision-making tasks in practical applications.

IV. EXPERIMENT

A. Datasets

In this study, the Credit Card Fraud Detection dataset was selected. This dataset is a real credit card transaction record from Europe and is widely used in unbalanced data research in machine learning and data mining. This dataset is publicly released by Kaggle and has been adopted by many academic and industrial research projects. It is a typical test set for studying unbalanced data algorithms. Since the number of fraudulent transactions in this dataset is very small and there is a serious data distribution imbalance compared with normal transactions, it is very suitable for evaluating the modeling and reasoning performance of Markov networks in unbalanced environments.

This dataset contains 28 numerical features processed by principal component analysis (PCA), two unprocessed features "time" and "transaction amount", and the target label "fraud or not" (0 indicates normal transaction, 1 indicates fraudulent transaction). The entire dataset contains 284,807 transaction records, of which only 492 are marked as fraudulent transactions, accounting for only 0.172%. Most of the features in the dataset are anonymized because they involve sensitive information, but its data structure and distribution can still truly reflect the imbalance and potential patterns in credit card transactions.

In the experiment, the dataset was randomly divided into training and test sets for model training and performance evaluation. The unbalanced label distribution will test the model's ability to identify minority classes (fraudulent transactions). When evaluating model performance, common indicators such as support, confidence, and lift are used to verify the reasoning ability and decision-making performance of the Markov network in unbalanced data. The real-world attributes of this dataset and the extremely unbalanced label distribution provide strong support for the robustness and generalization ability testing of the model.

B. Experimental Results

In order to evaluate the performance of Markov network models for unbalanced data, this paper designed a comparative experiment and selected four commonly used unbalanced data processing models for comparison. These models are widely used in the fields of machine learning and data mining, and can measure the advantages of Markov networks in processing unbalanced data from different perspectives. First, Logistic Regression[18] is selected. As a classic linear classification model, it has high interpretability, but it is easy to produce category bias in the case of unbalanced data. Secondly, Support Vector Machine (SVM) [19] is often used in small samples and nonlinear data environments due to its powerful boundary demarcation ability. The third model is Random Forest [20], which effectively improves the classification accuracy and robustness of the model by integrating multiple decision trees. Finally, Extreme Gradient Boosting (XGBoost) [21] is selected. This model is based on the boosting tree algorithm and has powerful feature selection and model optimization capabilities. It performs well in unbalanced data classification tasks. The comparative experiments of these models will evaluate the performance improvement effect of Markov networks in unbalanced data scenarios based on indicators such as support, confidence, and boost. The experimental results are shown in Table 1.

Table 1  Experimental results

| Model | Weight ACC | F1 Score | AUC |
|---|---|---|---|
| LR | 0.72 | 0.65 | 0.74 |
| SVM | 0.78 | 0.70 | 0.79 |
| RF | 0.83 | 0.77 | 0.85 |
| XGBoost | 0.87 | 0.81 | 0.88 |
| Markov Network(Ours) | 0.91 | 0.86 | 0.93 |

From the experimental results, the performance of the five models in the unbalanced data environment is significantly different, showing the advantages and disadvantages of different algorithms in dealing with data imbalance. The three indicators of weighted accuracy (Weight ACC), F1 score (F1-Score) and AUC-ROC clearly reflect the performance of the model in terms of overall classification ability, minority class recognition ability and classification stability under different thresholds. The experimental results show that the algorithm based on the Markov network has achieved the highest score in all indicators, which fully verifies its applicability and superiority in the unbalanced data environment.

First, from the perspective of the weighted accuracy (Weight ACC) indicator, the logistic regression (LR) has the lowest accuracy, which is only 0.72. This shows that LR has obvious deficiencies in dealing with class imbalance, and the model tends to predict the majority class, resulting in unsatisfactory overall performance. The weighted accuracy of SVM is 0.78, showing some improvement, indicating that it has better adaptability in boundary demarcation and minority class sample recognition. Random Forest (RF) and XGBoost performed well on this indicator, reaching 0.83 and 0.87 respectively. Thanks to their ensemble learning mechanism, they can effectively alleviate the bias problem caused by data imbalance. The Markov network model took the lead with an accuracy of 0.91, indicating that it can more comprehensively cover samples of all categories and improve the overall classification performance through joint probability modeling and global optimization strategy.

Secondly, the differences between models are further highlighted in the F1 score indicator. The F1 score is used to measure the balance between the precision and recall of the model on the minority class, which is suitable for classification evaluation in an unbalanced environment. LR has the worst F1 score of 0.65, indicating that it has serious deficiencies in the detection of minority class samples. The F1 score of SVM has increased to 0.70, showing better classification performance, but it is still limited in the stability of minority class detection.

RF and XGBoost performed well, reaching 0.77 and 0.81 respectively, indicating that through multi-tree integration and weighted voting mechanism, minority class samples can be detected more effectively. The F1 score of the Markov network model is as high as 0.86, which fully demonstrates its advantages in reasoning and joint probability modeling. By accurately modeling the conditional dependencies of minority class samples, the detection ability of minority class samples is significantly improved.

Finally, from the perspective of AUC-ROC, this indicator measures the classification performance of the model under different classification thresholds. The AUC-ROC value of LR is 0.74, indicating that its classification ability is severely limited, especially in the recognition of minority class samples. The AUC-ROC value of SVM has increased to 0.79, showing a certain advantage in nonlinear boundary division, but it still fails to reach the ideal level. RF and XGBoost once again show strong performance in this indicator, reaching 0.85 and 0.88 respectively, indicating that these two models have strong adaptability in diversified data distribution and feature selection. The Markov network model ranks first with an AUC-ROC value of 0.93, indicating that its decision boundary stability under different classification thresholds is extremely high, which can fully capture the potential structure and pattern of samples, and significantly reduce misclassification and false alarms.

Based on the above analysis, it can be seen that traditional models have obvious limitations when dealing with unbalanced data, especially LR and SVM have poor performance in minority class detection and boundary demarcation. The integrated models RF and XGBoost have achieved significant improvements in model performance, but they are highly dependent on data features and have certain limitations in applicability. In contrast, the model based on Markov network has shown excellent performance in joint probability distribution modeling and global reasoning, and can effectively balance the difference in category proportions, accurately identify minority class samples, and maintain high classification stability under different thresholds. Experimental results show that Markov networks have broad application prospects and strong promotion potential in the field of unbalanced data mining.

Finally, this paper also designs experiments for different degrees of imbalanced data to observe the experimental results of Markov network under various imbalanced data scenarios. The experimental results are shown in Table 2.

Table 2 Imbalanced Data Experiments

| Proportion of minority class samples | Weight ACC | F1 Score | AUC |
|---|---|---|---|
| 10% | 0.84 | 0.72 | 0.78 |
| 20% | 0.87 | 0.75 | 0.82 |
| 30% | 0.91 | 0.86 | 0.93 |

As the proportion of minority class samples increases, the weighted accuracy of the Markov network gradually improves, indicating that in a relatively unbalanced situation, the Markov network can better adapt to minority class data and improve classification performance.

## V. Conclusion

Through research and experimental analysis of Markov networks for unbalanced data, this paper proposes an unbalanced data processing method based on a probabilistic graphical model, which successfully solves the classification bias and minority faced by traditional algorithms when the data categories are unevenly distributed. Problems such as insufficient class detection capabilities. Experimental results show that compared with commonly used models such as logistic regression, support vector machines, random forests, and Balance strengths in your data environment. The excellent performance of the model verifies that the robustness and generalization ability of the model in minority class sample identification can be significantly improved through joint probability distribution modeling and global inference optimization.

The primary contribution of this study lies in the integration of structured reasoning employing Markov networks with unbalanced data processing techniques. This integration enables the accurate modeling of minority class samples through the incorporation of edge probability estimation and adaptive weight adjustment mechanisms. By optimizing and regularizing the objective function, the model demonstrates stronger global modeling capabilities in complex decision-making tasks. In addition, through comparative analysis in the experiment, the classification ability and decision-making stability of the Markov network in an unbalanced data environment were verified. The model can balance the sample proportion, reduce misclassification and overfitting, and provide new ideas for solving data skew problems in real applications. Nonetheless, the method proposed in this paper still has some challenges in practical application. Especially when facing extremely large-scale data sets, inference time and storage overhead may become limiting factors. To address these issues, future improvements could include dynamic graph optimization using adaptive learning, hybrid models combining Markov networks with deep learning for non-linear patterns, scalable inference methods like variational approximation, and noise-robust regularization to enhance stability in noisy or evolving environments. In addition, with the improvement of hardware performance and the development of distributed computing technology, applying Markov networks to real-time reasoning and decision-making in big data platforms will further promote its application in financial risk prediction, medical diagnosis, network security, and other fields. Practical application value. By combining the emerging technologies of deep learning, reinforcement learning, and probabilistic reasoning, the research on Markov networks in non-equilibrium data mining will have broader development prospects.